\documentclass[10pt,twocolumn,letterpaper]{article}
\usepackage{wacv}
\usepackage{times}
\usepackage{epsfig}
\usepackage{graphicx}
\usepackage{amsmath}
\usepackage{amssymb}
\usepackage{booktabs}
\usepackage{bbm}
\usepackage[accsupp]{axessibility}
\usepackage{multirow}

\newcommand{\figref}[1]{Fig.~\ref{#1}}
\newcommand{\tabref}[1]{Tab~\ref{#1}}
\newcommand{\figureref}[1]{Figure~\ref{#1}}
\newcommand{\tableref}[1]{Table~\ref{#1}}

\wacvfinalcopy 
\ifwacvfinal
\usepackage[breaklinks=true,bookmarks=false]{hyperref}
\else
\usepackage[pagebackref=true,breaklinks=true,colorlinks,bookmarks=false]{hyperref}
\fi

\begin{document}

\title{\textit{Self-Pair}: Synthesizing Changes from Single Source for Object Change Detection in Remote Sensing Imagery}

\author{Minseok Seo, Hakjin Lee, Yongjin Jeon, Junghoon Seo\\
SI Analytics\\
70, Yuseong-daero 1689beon-gil, Yuseong-gu,
Daejeon, Republic of Korea\\
{\tt\small \{minseok.seo, hakjinlee, yongjin117, jhseo\}@si-analytics.ai}
}

\maketitle
\thispagestyle{empty}

%

%
\begin{abstract}
For change detection in remote sensing, constructing a training dataset for deep learning models is difficult due to the requirements of bi-temporal supervision.
To overcome this issue, single-temporal supervision which treats change labels as the difference of two semantic masks has been proposed.
This novel method trains a change detector using two spatially unrelated images with corresponding semantic labels such as building.
However, training on unpaired datasets could confuse the change detector in the case of pixels that are labeled unchanged but are visually significantly different.
In order to maintain the visual similarity in unchanged area, in this paper, we emphasize that the change originates from the source image and show that manipulating the source image as an after-image is crucial to the performance of change detection.
Extensive experiments demonstrate the importance of maintaining visual information between pre- and post-event images, and our method outperforms existing methods based on single-temporal supervision. code is available at \url{https://github.com/seominseok0429/Self-Pair-for-Change-Detection}
\end{abstract}

\section{Introduction}
Change detection aims to detect the location of interest regions among semantically changed areas.
Generally, this change of interest (CoI) between multi-temporal high spatial resolution (HSR) remote sensing images is defined in the same area but at different times.
%
%
%

Recently, several supervised change detection methods~\cite{chen2021remote,fang2021snunet,ji2018fully,chen2022rdp} have been proposed and showed promising results.
%
Those methods are trained on the datasets consisting of pairs of bi-temporal images with change labels.
However, these bi-temporal datasets require high cost to obtain compared to the other tasks such as segmentation~\cite{waqas2019isaid} and detection~\cite{xia2018dota}.
%
It is due to several requirements for bi-temporal supervision.
First of all, obtaining correctly registered bi-temporal pair images is difficult due to the physical limitations of satellites.
Second, to decide whether the specific region is a change or not, observing both before- and after-image should be preceded.
Lastly, change is rare even in the real world; which makes it hard to obtain bi-temporal pair images containing change of interests.
For these reasons, publicly~\cite{benedek2009change, daudt2018urban, chen2020spatial, liu2020novel} opened change detection datasets are small-scaled and imbalanced.
%

To solve the problem, Zheng \textit{et al.}~\cite{zheng2021change} proposed a method to train a change detection model using only single-temporal labeled unpair images.
Instead of using bi-temporal labeled pair images, it trains the change detection model using a training dataset consisting of unpaired images which are randomly sampled from the training set.
\begin{figure}[t!]
    \centering
    \includegraphics[width=\columnwidth]{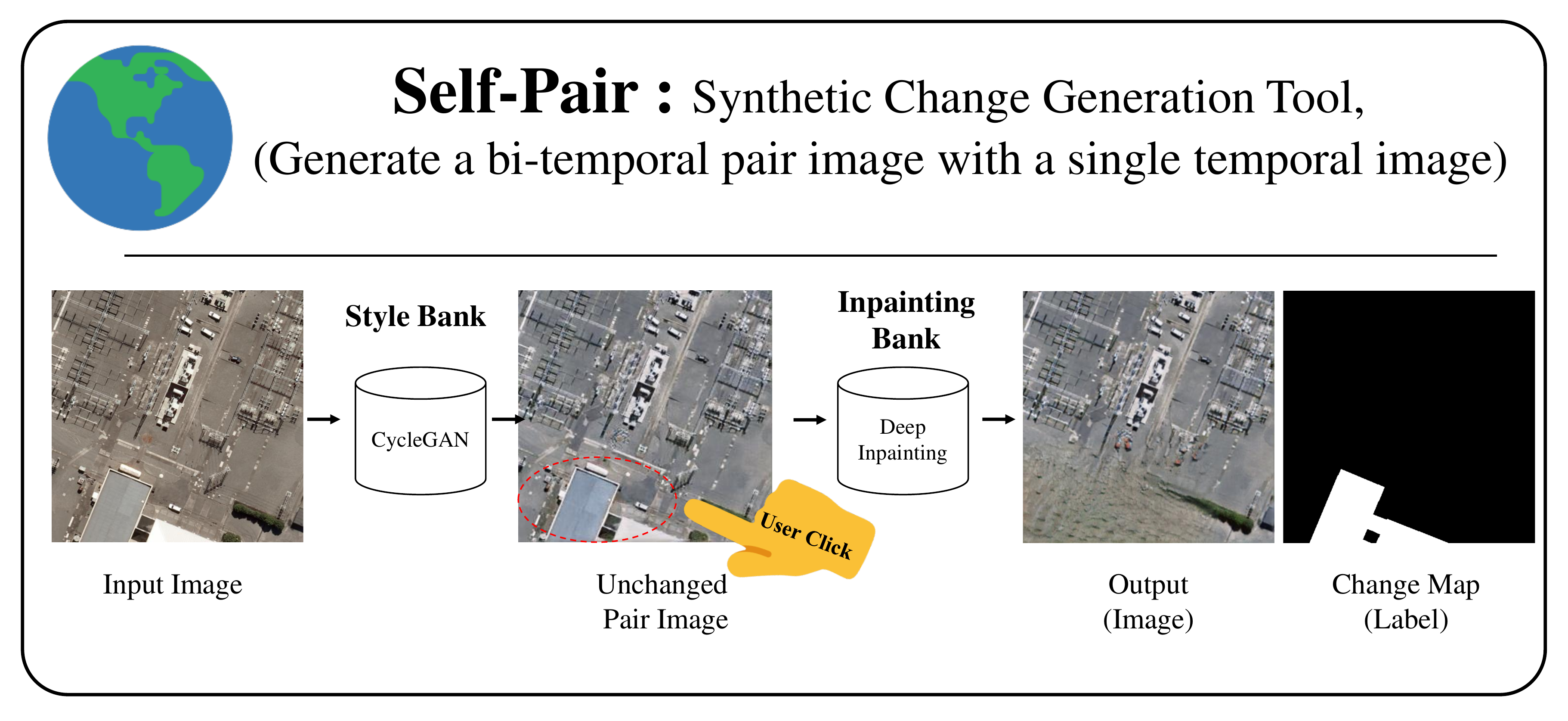}
    \caption{Synthetic change generation tool example.}
    \label{fig:teaser}
\end{figure}
The change label between two unpaired images is defined as whether the semantics of the same pixel location in two images are different or not (i.e.,\ \texttt{xor} operation).
This approach enables training change detectors without high-cost bi-temporal pairwise datasets, however, it ignores the structure and style consistency in unchanged areas.
In this setting, for example, buildings in paired images but with different color or texture are labeled unchanged.
Also different semantics such as road and grassland are also labeled unchanged regions since they are not change of interest.
This inconsistency makes the model confused to learn what the change is.
Furthermore, unpaired setting ignores useful context information such as object sizes, common patterns, or styles of the area, which can be obtained from bi-temporal paired images.

In this paper, we propose \textit{Self-Pair}, a novel synthetic image generation method of constructing input pairs for change detection models.
The key idea of \textit{Self-Pair} is that more diverse and realistic pair images can be generated with single-temporal single images, retaining the characteristics of real-world settings.
\textit{Self-Pair} relieves complicated conditions of change detection datasets such as high cost labeling, registration, and preserves the characteristics of unchanged regions as shown in the \figref{fig:teaser}.
We verified the performance of the existing change detection methods and \textit{Self-Pair} in both in-domain and cross-domain settings.
On extensive benchmarks, our approach outperforms previous single-temporal supervision method \cite{zheng2021change}, and even bi-temporal supervision methods in some cases.
Moreover, by conducted experiments on SNUNet-CD~\cite{fang2021snunet}, BIT-CD~\cite{chen2021remote}, and ChangeStar~\cite{zheng2021change} architectures, our method shows applicability to the various change detection architectures.


%
\section{Related Works}
\subsection{Object Segmentation in Remote Sensing}
Semantic segmentation in remote sensing images is challenging because of significant scale variation, background complexity, and imbalance between background and foreground.
To address these challenges, Zheng \textit{et al.}~\cite{zheng2020foreground} proposed \textit{FarSeg} in the perspective of foreground modeling. They achieved high performance with a better trade-off between speed and accuracy compared with general semantic segmentation methods~\cite{chen2017deeplab,zhao2017pyramid}.
Li~\textit{et al.}~\cite{li2021pointflow} achieved state-of-the-art through affinity context modeling, which focuses on solving the background complexity and background-foreground imbalance problem.


\subsection{Object Change Detection in Remote Sensing}
Change detection has been studied along with the rising need of utilizing remote sensing images to find meaningful changes~\cite{singh1989review,hussain2013change}.
As deep learning progresses, the change detection methods based on deep learning also show promising performance~\cite{mou2017multitemporal,zheng2022changemask,fang2021snunet,chen2021remote}.
However, these methods require  well-defined datasets with bi-temporal supervision, and most public datasets are small-scale, hence these methods show poor performance on the real-world cases~\cite{shen2021s2looking, noh2022unsupervised}.
The major reason for this phenomenon is that changes are rare compared to non-changes in bi-temporal paired images~\cite{zheng2021change}, and there is a difficulty in collecting bi-temporal paired images.
For this reason, the progress of change detection was relatively slow compared with the other tasks.

Recently, various change detection benchmark datasets have been proposed to solve this problem, but they still suffer from a lack of data samples~\cite{daudt2018urban,liu2020novel,chen2020spatial,gong2015change}.
To address this issue, Zheng \textit{et al.}~\cite{zheng2021change} recently proposed ChangeStar that detects changes using single temporal unpaired images with pseudo labels.
ChangeStar significantly alleviated the training data collection problem in change detection.
It uses \texttt{xor} operation on the semantic segmentation labels of two single temporal unpaired images to make pseudo change label, and use pseudo change label to train the change detection model.
However, ChangeStar does not consider the style, texture and consistency information coming from the bi-temporal paired images, which significantly degrades performance compared with the bi-temporal paired change detection methods.


\begin{figure*}[t!]
    \centering
    \includegraphics[width=1.5\columnwidth]{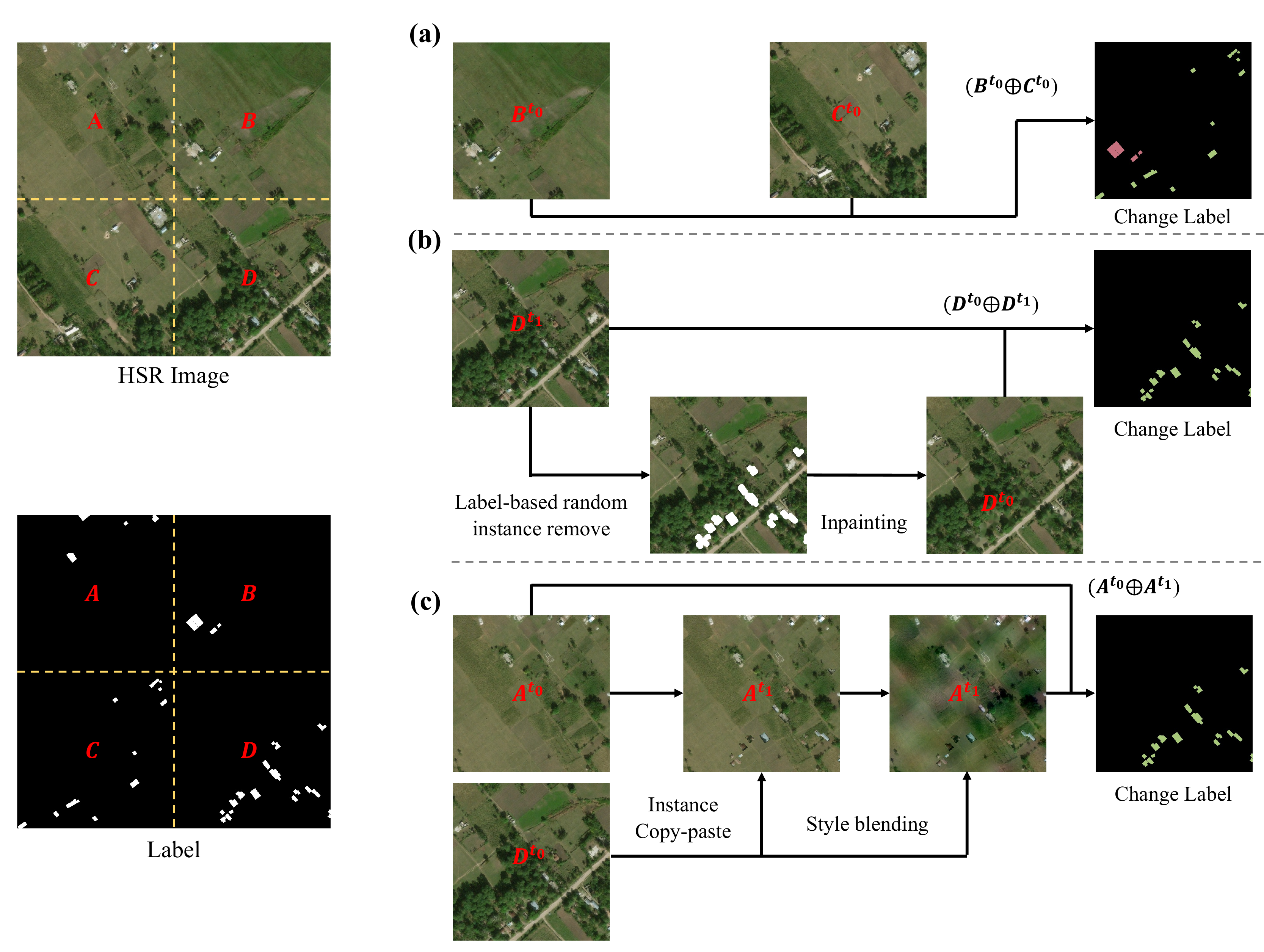}
    \caption{\textit{Self-Pair}. Three ways of generating single temporal paired images with pseudo change label using a single image. (a) Randomly cropping two patches from a single image. (b) Semantic label based inpainting approach. (c) Semantic label based copy-and-paste with style blending approach.}
    \label{fig:main}
\end{figure*}

\subsection{Data Augmentation in Remote Sensing Imagery}
In recent years, the strong augmentation strategy such as copy-and-paste (CP) and inpainting is widely used to various deep learning tasks(e.g. classification~\cite{zhang2017mixup,yun2019cutmix}, object detection~\cite{dvornik2018modeling}, and object segmentation~\cite{ghiasi2021simple}).
Also in remote sensing domain, there had been trial to exploit strong augmentation strategy. Kumdakcı \textit{et al.}~\cite{kumdakci2021generative} proposed inpainting method for augmenting the vehicle instances to solve the data shortage problem, however, it has limitations of generating instances with diversity.


\section{Approach}

\subsection{Change Detection from Bi-temporal Supervision}\label{sec:Re}
The goal of change detection based on bi-temporal supervision with given the pre-event images
$ \{\mathbf{X}_1^{t_0}, \dots, \mathbf{X}_N^{t_0} \vert$  $\mathbf{X}_k^{t_0} \in \mathbb{R}^{C \times H \times W}  \} $ and the corresponding post-event images $ \{\mathbf{X}_1^{t_1}, \dots, \mathbf{X}_N^{t_1} \vert \mathbf{X}_k^{t_1} \in \mathbb{R}^{C \times H \times W}  \} $ can be formulated as follows:
\begin{equation}
  \min_{\theta} \sum_{k=1}^{N} \mathcal{L}(\mathcal{F}_{\theta}(\mathbf{X}_k^{t_{0}}, \mathbf{X}_k^{t_{1}}), \mathbf{Y}_k^{t_{0} \rightarrow t_{1}} ),
  \label{eq:pair}
\end{equation}
\noindent where $\mathcal{L}$ indicates the loss function between the predicted change map obtained by change detector $\mathcal{F}_{\theta}$ with paired bi-temporal images $(\mathbf{X}_k^{t_{0}}, \mathbf{X}_k^{t_{1}})$, which indicates the pre-event image and the post-event image of a specific area $k$.
Change map, $\mathbf{Y}_k^{t_{0} \rightarrow t_{1}} \in \{0, 1\}^{H \times W}$ represents the regions of change of interest between the pre- and post-event images.

Deep learning change detection models based on bi-temporal paired images(\textit{Pair} training method), require the dataset to be a collection of image pairs.
This requirement leads high cost of building process and insufficient samples problem because set of image pairs should be taken from the same region at the different time and also should contain the CoIs.

\subsection{Disentangle the Change Map}
Zheng \textit{et al.}~\cite{zheng2021change} proposed ChangeStar, which defines change as an area where the same pixel position in pre- and post-event images has different semantic information.
From the definition of change by Zheng \textit{et al.}, we can disentangle the change map $\mathbf{Y}^{t_0 \rightarrow t_{1}}$ to the difference of the semantic information $\mathbf{Y}_k \oplus \mathbf{Y}_l$ from two different images $\mathbf{X}_k$ and $\mathbf{X}_l$.
The relaxed formulation without bi-temporal information $\{ \mathbf{X}_1^{t_1}, \dots, \mathbf{X}_N^{t_1} \}$ can be expressed as follows:
%
\begin{equation}
  \min_{\theta}  \sum_{k=1}^{N} \sum_{l=1}^{N} \mathbbm{1}(k \neq l) \mathcal{L}( \mathcal{F}_{\theta}(\mathbf{X}_k^{t_0}, \mathbf{X}_l^{t_0}), \mathbf{Y}_k^{t_0} \oplus \mathbf{Y}_l^{t_0} ),
  \label{eq:unpair}
\end{equation}
where $\mathbbm{1}$ and $\oplus$ denotes the indicator function and the ~\texttt{xor} operation, respectively.
Disentangle the change map alleviates the constraint in the original formulation by replacing paired images to unpaired images that having no geographical relationship. By disentangling the change map,  change detection model can be trained with pseudo bi-temporal paired images randomly sampled from semantic segmentation datasets. We will call this method the ~\textit{Unpair} training method.

\begin{figure*}[t!]
    \centering
    \includegraphics[width=1.9\columnwidth]{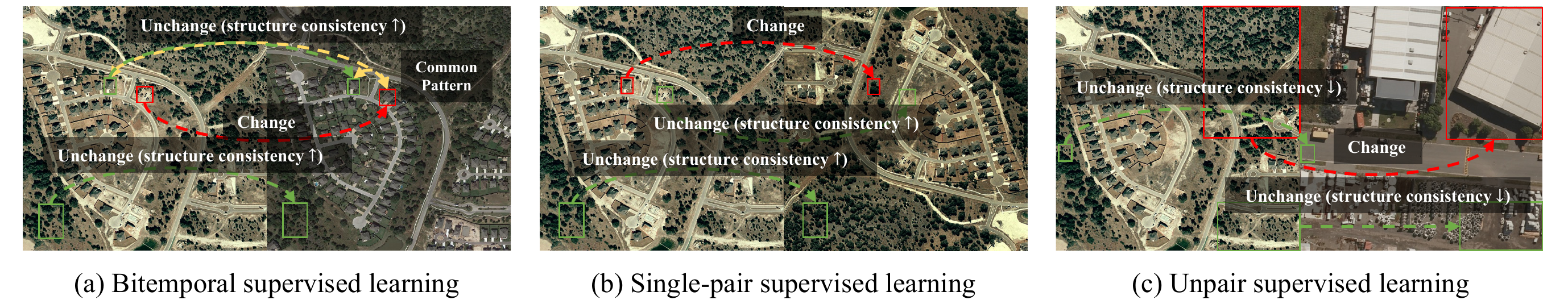}
    \caption{Qualitative comparison of structure similarity by bi-temporal, single-pair, and unpair scenarios.}
    \vspace{-2mm}
    \label{fig:Self-Pair_a}
\end{figure*}

~\textit{Unpair} training method achieved impressive performance with resolving the high cost problem of constructing bi-temporal paired dataset, however still having  a significant performance gap compared with  existing state-of-the-art methods that trained with bi-temporal supervision.
As shown in \figureref{fig:Self-Pair_a}-(a), ChangeStar ignores meaningful characteristics of bi-temporal paired images: consistency in unchanged area, common patterns, and style similarity across two images.
In~\figureref{fig:Self-Pair_a}-(c), areas in the green boxes which labeled as unchanged contain different semantic information (\textit{Left}: Tree, \textit{Right}: Road).
Such inconsistencies of unchanged regions across input images could confuse the model to learn what the change is, which leads to degradation of the performance. 
This indicates that the ChangeStar's \textit{Unpair} training formulation is missing essential properties of the change detection for the real-world bi-temporal setting.

\subsection{General Formulation of Change Detection}
To overcome the problems come from dataset and reflect the properties missed by \textit{Unpair} training formulation, we rethink how the change creates from given two pre- and post-event images: \textbf{all changes originate from the source and arise from the manipulation of the source}.
In order to represent this principle, we reformulate change detection as follows:
\begin{equation}
  \min_{\theta} \sum_{k=1}^{N} \mathcal{L}( \mathcal{F}_{\theta}(\mathbf{X}_k^{t_0}, g(\mathbf{X}_k^{t_0})), \mathbf{Y}_k^{t_0} \oplus g(\mathbf{Y}_k^{t_0}) ),
  \label{eq:general}
\end{equation}
where $g$ is a function that maps an image or label of specific area at particular time to an image or label of arbitrary time step where the change occurs.

This formulation allows us to utilize single-temporal  images $\mathbf{X}_k^{t_0}$ like ChangeStar, and to devise more plausible mapping functions $g$ that preserve meaning information in the real-world bi-temporal paired image setting.
With this formulation, in this paper, we focus on the study of designing the function $g$ and propose three simple and effective augmentation strategies.

\subsection{Proposed Method}
For the candidate of function $g$, photometric transformation and geometric transformation functions could be on the lists. However, only geometric transformation functions are discussed in this paper due to photometric transformation could not augment the new change instances. 
%
\vspace{-2mm}
\subsubsection{Random Crop from Single Image}

A very naive method for generating single-temporal paired images based on geometric transformation is random crop. Randomly cropped two patches with no overlapped regions from the source image are set as an input pair for change detection model.
This approach is similar to the method proposed in ~\cite{zheng2021change} but different in generating a pair from a single source rather than randomly sampled two images.
Although this strategy loses most of its structural consistency, it can still retain the style similarity or typical patterns which can be observed in the real-world bi-temporal paired images.
Moreover, this aggressive strategy could behave as a strong augmentation considering the massive changes that can happen in the real world as well.

For the experiment of random crop strategy, rotation operation to the cropped patches is mixed and comprehensive random crop method for change instance augmentation can be expressed as follows:
\begin{equation}
\begin{split}
  \min_{\theta} \sum_{k=1}^{N} \mathcal{L}(\mathcal{F}_{\theta}(\text{crop}_1(\mathbf{X}_k^{t_{0}}), r(\text{crop}_2(\mathbf{X}_k^{t_{0}})) ), \\ 
  \text{crop}_1(\mathbf{Y}_k^{t_{0}}) \oplus r(\text{crop}_2(\mathbf{Y}_k^{t_{0}})) ),
\end{split}
  \label{eq:crop}
\end{equation}

where $r$ is a random rotation function, $\text{crop}_1$ and $\text{crop}_2$ are notation of random crop function. Each crop functions crop same location for the corresponding image and label while cropping different locations for single-temporal paired setting.
%
\begin{figure*}[t!]
    \centering
    \includegraphics[width=1.8\columnwidth]{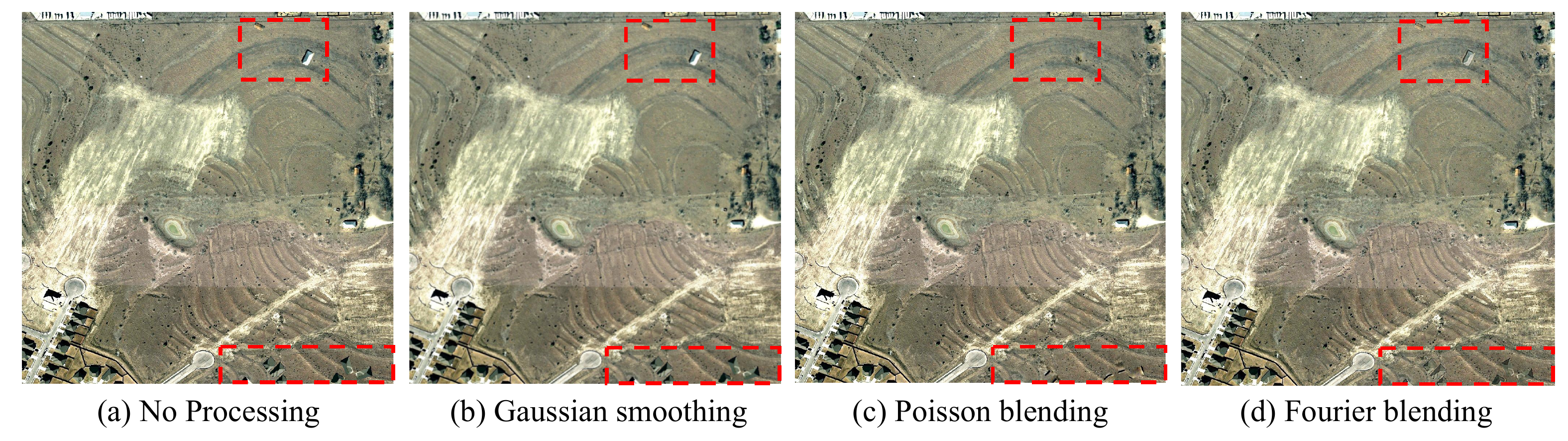}
    \caption{Qualitative comparison of blending methods used in Copy \& Paste~\cite{dvornik2018modeling} and our Fourier blending.}
    \label{fig:blending}
\end{figure*}
\subsubsection{Inpainting based on Labels}

One of the most common changes happens in the real world is the disappearance of objects.
We implement this behavior by erasing randomly selected instances and inpaint the background based on surroundings.
Since only minimal changes occur in the entire image, most structural consistency is maintained in the unchanged area.
Unlike random crop strategy, inpainting strategy preserves the important informations, such as structural consistency, common patterns, and style similarities that can be observed in bi-temporal paired settings.
Inpainting based change instance augmentation strategy can be expressed as below:
\begin{equation}
\begin{split}
  \min_{\theta} \sum_{k=1}^{N} \mathcal{L}(\mathcal{F}_{\theta}(\mathbf{X}_k^{t_{0}},  \text{inpaint}( (\mathbf{X}_k^{t_{0}} \times a), 1-a ),  \\ \mathbf{Y}_k^{t_{0}} \oplus (\mathbf{Y}_k^{t_{0}} \times a)),
\end{split}
  \label{eq:inpaint}
\end{equation}
where $a$ is binary mask of the objects to be erased, and these objects are randomly sampled at each time. 
For the implementation, we adopt method in \textit{Telea et.al}~\cite{telea2004image} for inpainting, and the inpainted images are set to pre-event images.

\subsubsection{Copy-and-Paste Instance Labels}
The other changes that commonly happens in the real world is occurance of the objects. 
A simple method for adding objects to an image is copying objects from the source image using semantic masks and pasting it to the target image.
Copy-and-Paste strategy can be used for easing the extreme imbalance between foreground and background that commonly observed in remote sensing images.

For the implementation of copy-and-paste strategy for augmenting the change instances, we copy objects from one of cropped patches and paste them to the other cropped patch.
Even though the objects are extracted from a single source, there may be an artifact near boundary of the pasted object due to the randomness of the paste location. To eliminate the artifacts to make augmented sample more realistic, fast Fourier transform based blending method is used.

Let $\mathcal{F}^{A}$, $\mathcal{F}^{P}$ 
be the amplitude and phase components of the Fourier transform $\mathcal{F}$,
\begin{equation}
 \mathcal{F}(x)(m,n) = \sum_{h,w}\mathbf{X}(h,w)e^{-j2\pi(\frac{h}{H}m+\frac{w}{W}n}), j^{2} = -1.
 \label{eq:fft}
\end{equation}
%
We denote $M_{\beta}$ as a mask for blending the amplitudes from each original image and modified image by copy-and-paste method. $M_{\beta}$ strictly followed the method proposed in~\cite{yang2020fda}.
%
%
%
Given two images (original image $\mathbf{X}_k^{t_{0}}$ and copy-and-pasted image $\mathbf{X}_k^{t_{0} \prime}$), realistically synthesized copy-and-pasted image using Fourier blending can be follw as:
\begin{equation}
\begin{split}
 \mathbf{X}_k^{t_{1} \prime} = \mathcal{F}^{-1}([M_{\beta}\circ\mathcal{F}^{A}(\mathbf{X}_k^{t_{0}})+(1-M_{\beta})\circ\mathcal{F}^{A}(\mathbf{X}_k^{t_{0} \prime}), \\ \mathcal{F}^{P}(\mathbf{X}_k^{t_{0} \prime})]).
\end{split}
 \label{eq:fft2}
\end{equation}

\figureref{fig:blending} is a visual comparison between the conventional blending method and our Fourier blending method.
Our Fourier blending method is inspired by ~\cite{yang2020fda,huang2021fsdr} that mix the styles of images from two different domains. However, unlike the those studies, we reduce the style gap at the modified regions by replacing the amplitude in the same image in the same domain.

\setlength{\tabcolsep}{4pt}
\renewcommand{\arraystretch}{1.5}
\begin{table*}[t!]
\centering
\resizebox{0.75\textwidth}{!}{%
\begin{tabular}{c|c|cccc|cccc|cccc}
\hline
\multirow{3}{*}{\textbf{Model}} & \multirow{3}{*}{\textbf{Method}} & \multicolumn{4}{c|}{\textbf{Train on xView2 pre-disaster}}                        & \multicolumn{4}{c|}{\textbf{Train on SpaceNet2}}                                  & \multicolumn{4}{c}{\textbf{Oracle}}                                              \\ \cline{3-14} 
                                &                         & \multicolumn{2}{c|}{WHU}                 & \multicolumn{2}{c|}{LEVIR-CD} & \multicolumn{2}{c|}{WHU}                 & \multicolumn{2}{c|}{LEVIR-CD} & \multicolumn{2}{c|}{WHU}                 & \multicolumn{2}{c}{LEVIR-CD} \\ \cline{3-14} 
                                &                         & IoU (\%) & \multicolumn{1}{c|}{F1 (\%)} & IoU (\%)      & F1 (\%)      & IoU (\%) & \multicolumn{1}{c|}{F1 (\%)} & IoU (\%)      & F1 (\%)      & IoU (\%) & \multicolumn{1}{c|}{F1 (\%)} & IoU (\%)      & F1 (\%)     \\ \hline
\multirow{3}{*}{SNUNet-CD}      & Pair                    &         & \multicolumn{1}{c|}{}          &              &                &         & \multicolumn{1}{c|}{}          &              &                & 74.54   & \multicolumn{1}{c|}{87.10}     & 81.93        & 92.11         \\
                                & Unpair                  & 64.28   & \multicolumn{1}{c|}{72.11}     & 71.22        & 80.05          & 66.91   & \multicolumn{1}{c|}{74.77}     & 64.01        & 70.98          &         & \multicolumn{1}{c|}{}          &              &               \\
                                & Self-Pair               & \textbf{69.19}   & \multicolumn{1}{c|}{\textbf{79.40}}     & \textbf{77.51}        & \textbf{84.52}          & \textbf{72.95}   & \multicolumn{1}{c|}{\textbf{81.70}}     & \textbf{69.38}        & \textbf{79.83}          &         & \multicolumn{1}{c|}{}          &              &               \\ \hline
\multirow{3}{*}{BIT-CD}         & Pair                    &         & \multicolumn{1}{c|}{}          &              &                &         & \multicolumn{1}{c|}{}          &              &                & 74.48   & \multicolumn{1}{c|}{86.07}     & 81.51        & 90.86         \\
                                & Unpair                  & 60.15   & \multicolumn{1}{c|}{70.22}     & 63.29        & 73.43          & 66.12   & \multicolumn{1}{c|}{73.85}     & 63.10        & 69.23          &         & \multicolumn{1}{c|}{}          &              &               \\
                                & Self-Pair               & \textbf{68.37}   & \multicolumn{1}{c|}{\textbf{78.76}}     & \textbf{72.91}        & \textbf{82.54}          & \textbf{71.81}   & \multicolumn{1}{c|}{\textbf{80.74}}     & \textbf{67.04}        & \textbf{77.29}          &         & \multicolumn{1}{c|}{}          &              &               \\ \hline
\multirow{3}{*}{ChangeStar}     & Pair                    &         & \multicolumn{1}{c|}{}          &              &                &         & \multicolumn{1}{c|}{}          &              &                & 79.89   & \multicolumn{1}{c|}{87.92}     & 91.09        & 94.91         \\
                                & Unpair                  & 75.61   & \multicolumn{1}{c|}{82.29}     & 80.13        & 88.65          & 64.51   & \multicolumn{1}{c|}{72.14}     & 60.20        & 68.32          &         & \multicolumn{1}{c|}{}          &              &               \\
                                & Self-Pair               & \textbf{82.94}   & \multicolumn{1}{c|}{\textbf{91.09}}     & \textbf{84.59}        & \textbf{93.14}          & \textbf{77.58}   & \multicolumn{1}{c|}{\textbf{83.90}}     & \textbf{74.41}        & \textbf{81.22}          &         & \multicolumn{1}{c|}{}          &              &               \\ \hline
\end{tabular}
}
\caption{Experimental results of Unpair and \textit{Self-Pair} methods in xView2 pre-disaster → (WHU, LEVIR-CD) and SpaceNet2 → (WHU, LEVIR-CD) cross domain tasks. Oracle is a single domain bi-temporal supervised training setup.}
\label{tab:main}
\end{table*}
\setlength{\tabcolsep}{1.4pt}

\section{Experiments}
 We evaluate our method in a cross-domain setting, that train the model with building segmentation dataset and validate under building change detection datasets which are constructed in a different purpose. For more fair comparison of \textit{Unpair, Pair,} and \textit{Self-Pair}, evaluation under in-domain setting is also conducted. 
Note that the LEVIR-CD dataset only offers change labels between $t_{0}$ and $t_{1}$ images, so in-domain experiments could not be performed.


%

\subsection{Experimental Settings}\label{sec:settting}
\noindent\textbf{Training Datasets.} Three building segmentation datasets for remote sensing are used to train change detectors in the formulations of \textit{Unpair} Eq.(\ref{eq:unpair}) and \textit{Self-Pair} Eq.(\ref{eq:general}), which exploit only single temporal supervision:
\begin{itemize}
    \item \textbf{xView2 pre-disaster}~\cite{zheng2021change}\textbf{.} xView2 dataset is originally proposed for building damage assessment. The pre-disaster dataset, which is a subset of the xView2 dataset contains 9,168 pre-disaster HSR images and 316,114 building polygons. We use subset from \textit{train} and \textit{tier3} split dataset. Each image has a size of 1,024 $\times$ 1,024 pixels.
    \item \textbf{SpaceNet2}~\cite{van2018spacenet}\textbf{.} SpaceNet2 dataset consists of 10,590 HSR images of size 650 $\times$ 650 pixels of 0.3 m GSD and 219,316 urban building instance annotations. Following ChangeStar, we also used only 3-bands pan-sharpened RGB images and their annotations.
    \item \textbf{WHU building change detection}~\cite{ji2018fully}\textbf{.} WHU dataset is constructed with one pair of aerial images of size 15,354 $\times$ 32,507 pixels obtained in 2012 and 2016 of the same area. It provides 12,796 and 16,077 building instance labels respectively, and change labels across the pair. Train, validation, and test set are composed of 4,736, 1,036, and 2,416 tiles that extracted from original aerial image pair.
\end{itemize}
\paragraph{Evaluation Datasets.} We evaluate \textit{Unpair} and \textit{Self-Pair} methods on the LEVIR-CD and WHU building change detection datasets, which are widely used in change detection evaluation.
\begin{itemize}
    \item \textbf{LEVIR-CD}~\cite{chen2020spatial}\textbf{.} The LEVIR-CD dataset contains 637 bi-temporal pairs of HSR images and 31,333 change labels on building instances. Each image has a size of 1,024 $\times$ 1,024 pixels with 0.5 m GSD. The change label provides the information about occurance of new buildings and disappearance of existing buildings. Train, validation and test set are split into 445, 128, and 64 pairs. Evaluation of \textit{Unpair} and \textit{Self-Pair} method are conducted with the test set.
\end{itemize}
\paragraph{Implementation details} We experiment our augmentation strategies based on three state-of-the-art change detectors{SNUNet-CD~\cite{fang2021snunet}, BIT-CD~\cite{chen2021remote}, and ChangeStar~\cite{zheng2021change}}.
Since these studies were conducted on different settings of backbone, optimizer and training schedule, experiments of \textit{Self-Pair} follow the most of hyperparameters of each studies' experimental details.
For SNUNet-CD, 16-channel model is adopted, and for BIT-CD and ChangeStar, ResNet50 backbone~\cite{he2016deep} is adopted.
Three ways of augmentation approaches of \textit{Self-Pair} are applied with the same probability in the training stage.
%
%
%
\subsection{Cross-domain Evaluation Results}\label{sec:cross}
For evaluating the \textit{Self-Pair},  two formerly proposed methods(\textit{Pair} and \text{Unpair}) are used for comparison. \textit{Pair} method trains the model with bitemporal supervision and evaluated under in-domain setting which can be considered as an upper bound.
Unlike \textit{Pair} and \textit{Unpair}, \textit{Self-pair} trains the model with single-temporal supervision and evaluated under cross-domain settings to checkout the generalization performance. Here \textit{Unpair} trains the models with Zheng \textit{et al.}~\cite{zheng2021change}'s  method, and \textit{Self-pair} indicates the models trained with our proposed method.

As shown in~\tableref{tab:main}, all of change detectors trained with our augmentation method outperform the change detectors trained with \textit{Unpair} method regardless of its architecture.
Even for the ChangeStar model, the performance of the model trained with our method outperforms the model trained with \textit{Pair} method.
This implies that \textit{Self-Pair} method can approximate the distribution of changes in the real-world better than given fixed dataset. 
\subsection{In-Domain Evaluation Results}
\label{sec:indomain}
We evaluate the performance of ChangeStar model trained with each \text{Pair}, \textit{Unpair}, and \textit{Self-Pair} on the WHU and LEVIR-CD dataset to compare the in-domain and cross-domain performance.
The experimental results are in ~\tabref{tab:ablation3}. As shown in ~\tabref{tab:ablation3}, our \textit{Self-Pair} method based ChangeStar achieved the best performance not only under in-domain experiment on the WHU dataset but also in the cross-domain experiment with LEVIR-CD dataset. In addition, the \textit{Pair} method showed the lowest performance in the cross-domain experiment, and was not significantly different from the \textit{Unpair} method even in the in-domain experiment. Those results indicate that paired images are not essential in both in-domain and cross-domain settings for change detection. \\
~\tableref{tab:ablation4} is an experiment to analyze the effect of each of three components in \textit{Self-Pair}. As shown in ~\tabref{tab:ablation4}, it can be seen as all components of \textit{Self-Pair} have a complementary relationship.
Also, when both of \textit{Self-Pair} and \textit{Pair} method are used for training, the model showed 3.34\% higher performance than trained with \textit{Self-Pair} method alone. Eventually, summarizing the results of ~\tabref{tab:ablation3} and ~\tabref{tab:ablation4}, it shows that paired input setting is not essential for change detection, while can give advantages for performance improvement.

\begin{figure*}[t!]
    \centering
    \includegraphics[width=1.6\columnwidth]{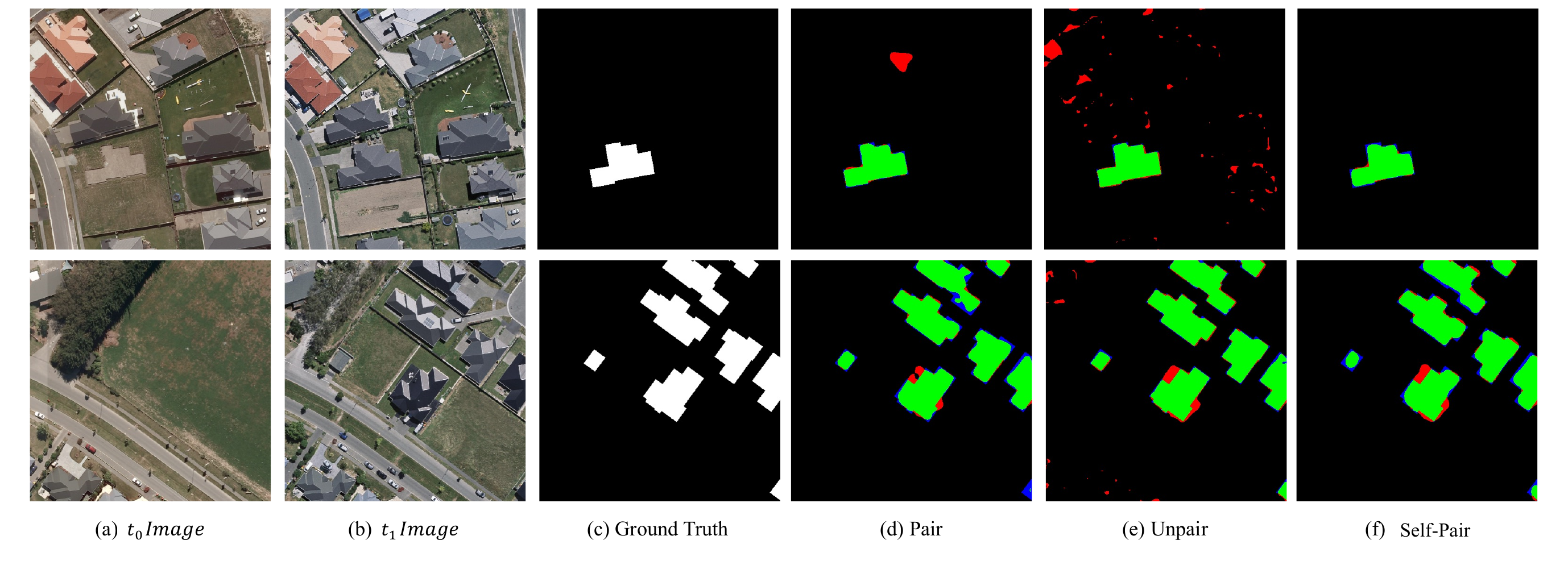}
    \caption{Qualitative analysis of pair, unpair, and Self-Pair. \textcolor{green}{True positives (TP)}, \textcolor{red}{false positives (FP)}, and \textcolor{blue}{false negatives (FN)} are represented as green, red, and blue, respectively.}
    \label{fig:vis}
\end{figure*}
\begin{table}[t!]
\centering
\resizebox{0.35\textwidth}{!}{%
\begin{tabular}{c|c|cccc}
\hline
           &           & \multicolumn{4}{c}{\textbf{Train on WHU  Trainset}}              \\ \cline{3-6} 
\textbf{Model}      & \textbf{Method}    & \multicolumn{2}{c}{WHU}               & \multicolumn{2}{c}{LEVIR-CD} \\ \cline{3-6} 
           &           & IoU (\%)  & \multicolumn{1}{c|}{F1 (\%)} & IoU (\%)        & F1 (\%)       \\ \hline
ChangeStar & Unpair    &    78.13      & \multicolumn{1}{c|}{86.41}      &              59.29  &         68.82    \\
ChangeStar & Self-Pair &    \textbf{83.57}      & \multicolumn{1}{c|}{\textbf{90.77}}      &   \textbf{66.79}              &     \textbf{78.41}        \\
ChangeStar & Pair      &    79.89      & \multicolumn{1}{c|}{87.92}      &      51.23          &    55.11        \\ \hline
\end{tabular}%
}
\caption{Evaluation result on in-domain (WHU Testset), cross-domain (LEVIR-CD Testset) performance according to each augmentation method(\textit{Pair, Unpair} and \textit{Self-Pair}).}
\label{tab:ablation3}
\vspace{-3mm}
\end{table}

\subsection{Qualitative Results}\label{sec:qual}
~\figureref{fig:vis} shows the results of qualitative analysis of camparison between \textit{Pair, Unpair}, and \textit{Self-Pair} on the WHU building change detection dataset.
As shown in \figref{fig:vis}-(e), \textit{Unpair} shows high TP (True Positive) score in both examples but also shows high FP (False Positive) score as well.
Compare with \textit{Pair}, \textit{Self-Pair} shows lower FP and FN (False Negative) with showing higher TP.
To summarize, results of comparison of the qualitative analysis show that for a task which only focusing the true positive score, using \textit{Unpair} method is efficient, or for the task which requires lower false positives and higher true positives, using \textit{Self-Pair} could be a best augmentation mthod.

    


\subsection{Discussion and Ablation Study}\label{sec:able}
\noindent\textbf{Why \textit{Self-Pair} works?} \textit{Self-Pair} is a strategy to generate a visually plausible realistic synthetic image to consist single-temporal paired images from a single image source. If \textit{Self-Pair} could create the characteristics of the bi-temporal paired images, the domain gap between bi-temporal paired setting and the \textit{Self-Pair} setting should be small \cite{long2015learning}. 


%
\begin{figure*}[t!]
    \centering
    \includegraphics[width=1.9\columnwidth]{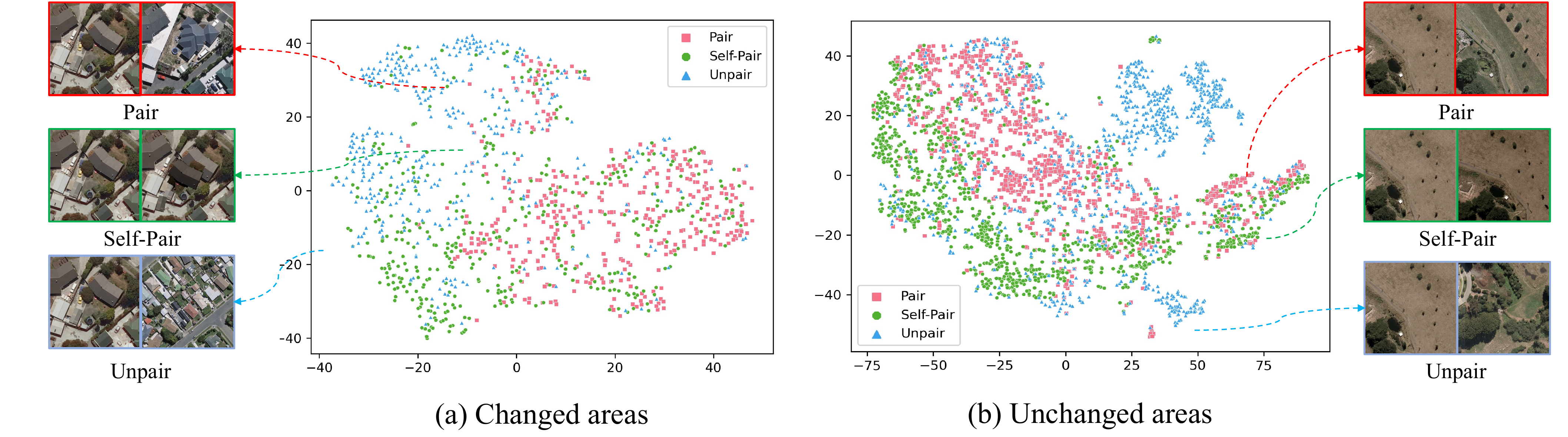}
    \caption{The result of t-SNE~\cite{van2008visualizing} by concatenating the intermediate features of pre and post-event images from ChangeStar which trained on the WHU change detection dataset. (a) is the result of the area where the change occurred, and (b) is the result of the area where having no changes.}
    
    \label{fig:tsne}
\end{figure*}

\begin{table}
\centering
\resizebox{0.33\textwidth}{!}{%
\begin{tabular}{c|ccccc|cc|cc}
\hline
\textbf{Method} & \multicolumn{5}{c|}{\textbf{Components}} & \multicolumn{2}{c|}{\textbf{Metric}} & \multicolumn{2}{c}{\textbf{Gain}} \\ \cline{2-10} 
                & UN     & CR     & IN     & CP    & PA    & IoU               & F1               & IoU              & F1             \\ \hline
Unpair          & \checkmark       &         &        &       &       &              78.13     &            86.41      &   0               &         0       \\
Self-Pair       &        & \checkmark       &        &       &       &               79.23    &  87.10                &  +1.10                &        +0.69        \\
Pair            &        &        &        &       & \checkmark      &               79.89     &         87.92         &          +1.76        &   +1.51             \\ \hline
Self-Pair       &        & \checkmark       & \checkmark       &       &       &               82.96    &   90.21               &             +4.83     &     +3.80           \\
Self-Pair       &        & \checkmark       & \checkmark       & \checkmark   &      &           83.57       &        90.77           &   +5.44               &  +4.36              \\ \hline
Pair            &        & \checkmark      & \checkmark        &    \checkmark   & \checkmark      &                  86.91 &         94.02    &        +8.78          &   +7.61             \\ \hline
\end{tabular}%
}
\caption{The result of comparing the effect of each component of \textit{Self-Pair} (\textit{UN}: \textit{Unpair}, \textit{CR}: Crop-and-Rotation, \textit{IN}: Inpainting, \textit{CP}: Copy-and-Paste with Blending, \textit{PA}: \textit{Pair})
.}
\label{tab:ablation4}
\end{table}

%
%
~\figureref{fig:tsne} shows the results of t-SNE embedding \cite{van2008visualizing} of \textit{Self-Pair}, bi-temporal paired setting (\textit{Pair}), and single-temporal paired setting (\textit{Unpair}). As show in ~\figureref{fig:tsne}, samples from \textit{Self-Pair} embed near to the embedded samples from \textit{Pair} in both cases where the change occurred and no changes. This indicates that \textit{Self-Pair} creates the synthetic images by utilizing characteristics of bi-temporal paired images. Note that \textit{Self-Pair} can be embedded more widely than embedding result in ~\figref{fig:tsne}, during training phase by randomness from inpainting method and copy-paste with blending method.
~\tableref{tab:emd} shows Earth mover's distance (EMD) \cite{bonneel2011displacement} between sets of intermediate features. 
Regardless of whether sample is labeled to change or non-change, $EMD \text{ } (\textit{Self-Pair}, \textit{Pair})$ costs smaller distance than $EMD \text{ } (\textit{Unpair}, \textit{Pair})$.
This tendency is also maintained regardless of whether the cosine metric or euclidean metric is used for compute the distance.
Accordingly, the results quantitatively proves that our method effectively reduces the domain gap better than the existing single-temporal paired strategy\cite{zheng2021change}.

\setlength{\tabcolsep}{4pt}
\begin{table}[t!]
\centering
\resizebox{0.33\textwidth}{!}{%
    \begin{tabular}{ccc}
    \hline
    \multicolumn{1}{c|}{\textbf{Cost}}                                                                  & \textbf{EMD(Pair, Self-Pair)} & \textbf{EMD(Pair, Unpair)} \\ \hline
    \multicolumn{3}{c}{(a) \textit{Changed Areas}}                                                                                                            \\ \hline
    \multicolumn{1}{c|}{\begin{tabular}[c]{@{}c@{}}Cosine\\[-9pt] \scriptsize($\times 10^{-2}$)\end{tabular}} & \textbf{4.3173}               & 10.6672           \\
    \multicolumn{1}{c|}{\begin{tabular}[c]{@{}c@{}}Euclidean\\[-9pt] \scriptsize($\times 1$)\end{tabular}} & \textbf{2.3946}               & 3.7833            \\ \hline
    \multicolumn{3}{c}{(b) \textit{Unchanged Areas}}                                                                                                          \\ \hline
    \multicolumn{1}{c|}{\begin{tabular}[c]{@{}c@{}}Cosine\\[-9pt] \scriptsize($\times 10^{-2}$)\end{tabular}} & \textbf{3.4424}               & 5.9130            \\
    \multicolumn{1}{c|}{\begin{tabular}[c]{@{}c@{}}Euclidean\\[-9pt] \scriptsize($\times 1$)\end{tabular}} & \textbf{1.5830}               & 2.2914            \\ \hline
    \end{tabular}
}
\caption{Earth mover's distance (EMD) between sets of intermediate features from \{\textit{Pair} and \textit{Self-Pair}\} or \{\textit{Pair} and \textit{Unpair}\} shown in ~\figref{fig:tsne}. \textit{Cost} indicates which type of distance metric is used.}
\label{tab:emd}
\end{table}
\begin{table}
\centering
\resizebox{0.33\textwidth}{!}{%
\begin{tabular}{c|c|c|c}
\hline
\textbf{Model}  & \textbf{Method}   & \textbf{IoU} & \textbf{Gain} \\ \hline
FarSeg~\cite{zheng2020foreground} & Baseline &   71.50  &    0  \\
FarSeg~\cite{zheng2020foreground} & Naive-CP~\cite{ghiasi2021simple} &   69.29  &-2.21      \\
Farseg~\cite{zheng2020foreground} & CPwB  &   74.71  &    +3.21  \\ \hline
PFNet~\cite{li2021pointflow}  & Baseline &   70.98  &  0    \\
PFNet~\cite{li2021pointflow}  & Navie-CP~\cite{ghiasi2021simple} &  69.10   & -1.88     \\
PFNet~\cite{li2021pointflow}  & CPwB  &  71.91   &  +0.93    \\ \hline
\end{tabular}%
}
\caption{Evaluation result of building semantic segmentation according to each
naive copy-and-paste and copy-and-paste with blending augmentation strategy.
}
\label{tab:ablation1}
\end{table}

\begin{figure}[t!]
    \centering
    \includegraphics[width=0.6\columnwidth]{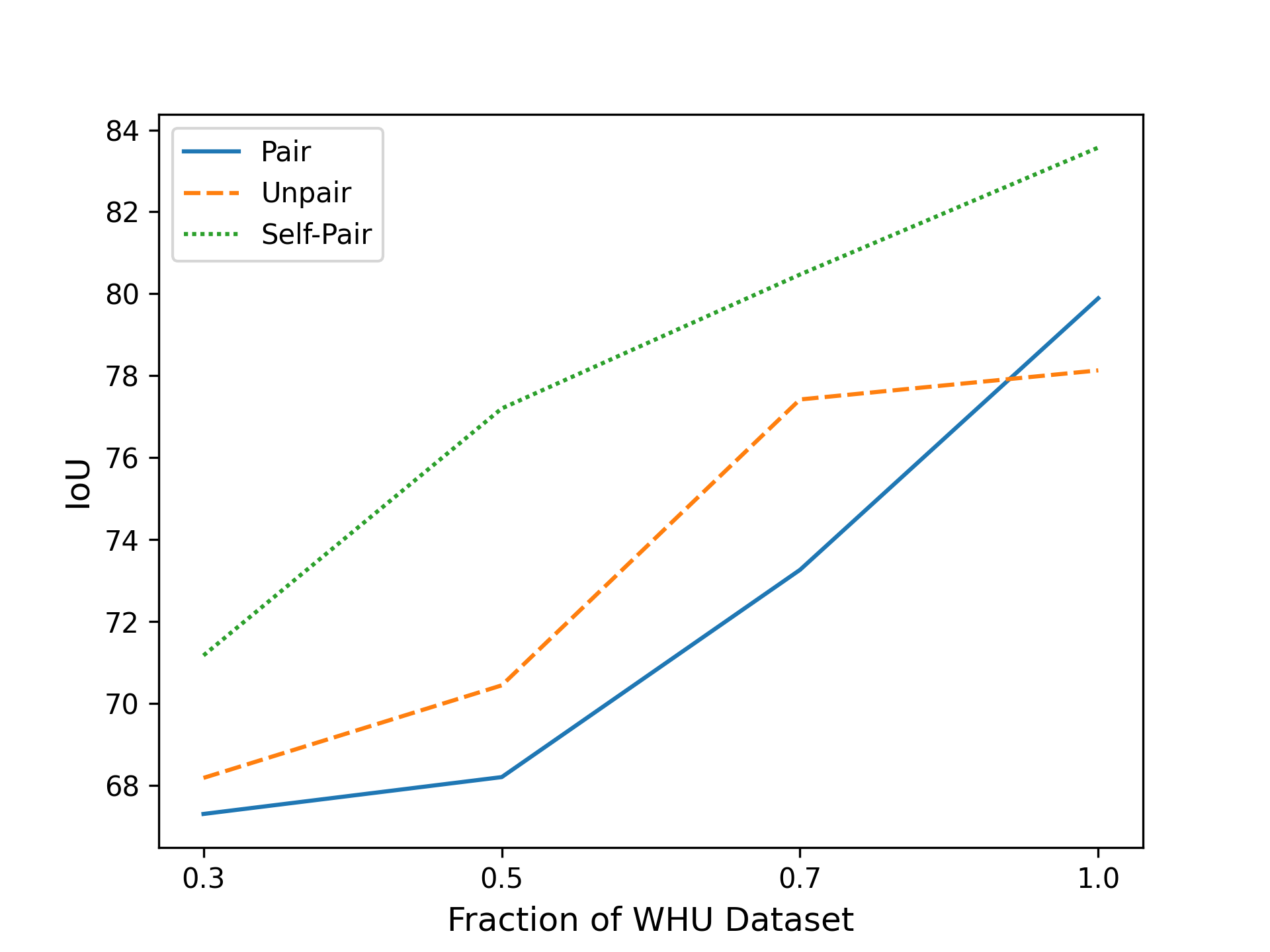}
    \caption{Data-efficiency on the WHU benchmark. Note that \textit{Self-Pair} and \textit{Unpair} requires three times as many epochs as pairs for convergence.}
    \label{fig:data}
\end{figure}

\noindent\textbf{Ablation study} \tableref{tab:ablation1} shows the results of comparing naive copy-and-paste (Naive-CP)~\cite{ghiasi2021simple} and copy-and-paste with blending (CPwB) in building semantic segmentation dataset.
As shown in the table, performance of Naive-CP is lower than baseline.
For the semantic segmentation on the remote sensing domain, background context is highly important(e.g.\ no building on the sea). Also, scale of objects are having difference between different scenes.
This is why the performance of Naive-CP that designed without considering characteristics of HSR remote sensing images is significantly reduced.
On the other hand, CPwB is using only one single image for setting an input pair for change detection model which means that there is no scale variation and the context of background is fixed.
For these reasons, CPwB could easily adapt to the characteristics of remote sensing imagery domain and showing improved results to the baseline while Naive-CP fails to adapt to it and shows degraded results.

~\tableref{tab:ablation2} shows the result of comparing the performance of Gaussian smoothing, Poisson blending~\cite{perez2003poisson} which used in~\cite{dvornik2018modeling}, and our Fourier blending.
Gaussian smoothing has no improvements and Poisson blending performed poorly compared with copy-and-paste with no blending. To analyze this phenomenon, we visualize the augmented samples based on each blending methods in ~\figref{fig:blending}.
In ~\figref{fig:blending}, the pasted buildings shown in the red boxes are very small and in the ~\figref{fig:blending}-(c), most of pasted buildings are erased after blended with poisson method. 
However, in ~\figref{fig:blending}-(d) every pasted buildings are left with changed texture when fourier method is used for blending.
Consequently, qualitative and quantitative results shows that blending method can mitigate the artifacts which created by hard augmentation and leads to the performance improvements.

%
\noindent\textbf{Data-efficiency on the WHU dataset.} \textit{Self-Pair} augments data in various forms according to three strategies.
~\figureref{fig:data} shows the results of evaluating the data efficiency of \textit{Pair}, \textit{Unpair}, and \textit{Self-Pair}.
\textit{Self-Pair} significantly improves data efficiency of the WHU building change detection dataset.

\begin{table}
    \centering
    \resizebox{0.3\textwidth}{!}{%
    \begin{tabular}{c|c|c|c}
    \hline
    \textbf{Model}  & \textbf{Method}             & \textbf{IoU} & \textbf{Gain} \\ \hline
    FarSeg~\cite{zheng2020foreground} & Gaussian smoothing &74.54     & -0.17      \\
    FarSeg~\cite{zheng2020foreground} & Poisson blending~\cite{perez2003poisson}   &    68.90 &  -2.60    \\
    Farseg~\cite{zheng2020foreground} & Fourier blending    &   76.19  & +1.48     \\ \hline
    PFNet~\cite{li2021pointflow}  & Gaussian smoothing &  72.29   & +0.38     \\
    PFNet~\cite{li2021pointflow}  & Poisson blending~\cite{perez2003poisson}   & 69.78     &  -2.13    \\
    PFNet~\cite{li2021pointflow}  & Fourier blending    &   73.41  &  +1.50    \\ \hline
    \end{tabular}%
    }
    \caption{The result of comparing the effect of each blending methods.}
    \label{tab:ablation2}
\end{table}

\section{Conclusions}
In this work, we redefine the change detection problem in way of how the change happens  - \textit{how to model the changes happen in the real-world}.
We proposed a novel data augmentation method \textit{Self-Pair}, which generates the synthetic image for constructing a input pair based on single-temporal single image and alleviates the problem of high cost of collecting pair set which contains changes for bi-temporal paired supervised learning.
%
%
%
We hope our method reduces the time cost of data collection and makes object change detection research more accessible, scalable, and economical.


\section*{Acknowledgement}
This work was supported by Artificial intelligence industrial convergence cluster development project funded by the Ministry of Science and ICT (MSIT, Korea) \& Gwangju Metropolitan City.
{\small
\bibliographystyle{ieee_fullname}
\bibliography{egbib}
}

\end{document}